\documentclass[10pt,journal,compsoc]{IEEEtran}
\usepackage{amsmath,amsfonts}
\usepackage{algorithmic}
\usepackage{array}
\usepackage[caption=false,font=normalsize,labelfont=sf,textfont=sf]{subfig}
\usepackage{textcomp}
\usepackage{stfloats}
\usepackage{url}
\usepackage{verbatim}
\usepackage{graphicx}

\usepackage{balance}
\usepackage{tikz}
\usepackage{comment}
\usepackage{amsmath,amssymb} 
\usepackage{color}

\usepackage{booktabs}

\usepackage{multicol} 
\usepackage{multirow} 

\newcommand\fig{Fig.}
\newcommand\tab{TABLE.}

\begin{document}

\title{Exploring the Impacts from Datasets to Monocular Depth Estimation (MDE) Models with MineNavi}

\author{
  Wang~Xiangtong,~
  Liang~Binbin,~
  Yang~Menglong,~
  and~Li~Wei,~

  \IEEEcompsocitemizethanks{
    \IEEEcompsocthanksitem Li~Wei was with the School of Astronautics and Astronautics, Sichuan University, Chengdu, China, 610207.\protect\\

    E-mail: li.wei@scu.edu.cn
    

  }

  \thanks{
    Manuscript received xx xx, xxxx; revised xx xx, xxxx.
  }

}

\markboth{
  Journal of \LaTeX\ Class Files,~Vol.~xx, No.~xx, xx~xxxx
  }%
{Shell 
\MakeLowercase{\textit{et al.}}: Bare Demo of IEEEtran.cls for Computer Society Journals
}

\IEEEtitleabstractindextext{%

\begin{abstract}
    Current computer vision tasks based on deep learning require a huge amount of data with annotations for model training or testing, especially in some dense estimation tasks, such as optical flow segmentation and depth estimation. In practice, manual labeling for dense estimation tasks is very difficult or even impossible, and the scenes of the dataset are often restricted to a small range, which dramatically limits the development of the community. To overcome this deficiency, we propose a synthetic dataset generation method to obtain the expandable dataset without burdensome manual workforce. By this method, we construct a dataset called MineNavi containing video footages from first-perspective-view of the aircraft matched with accurate ground truth for depth estimation in aircraft navigation application. We also provide quantitative experiments to prove that pre-training via our MineNavi dataset can improve the performance of depth estimation model and speed up the convergence of the model on real scene data. Since the synthetic dataset has a similar effect to the real-world dataset in the training process of deep model, we also provide additional experiments with monocular depth estimation method to demonstrate the impact of various factors in our dataset such as lighting conditions and motion mode. 
\end{abstract}

}

\maketitle

\IEEEdisplaynontitleabstractindextext

\IEEEpeerreviewmaketitle


\IEEEraisesectionheading{
    \section{Introduction}
    \label{sec:introduction}
}

\IEEEPARstart{I}{n} recent years, the machine learning based depth estimation methods, which heavily rely on the labeled dataset, have achieved satisfying performance. However, the scarcity of available labeled data, high costs of data acquisition and annotation, limit the quantity and variety of existing deep learning methods.
Although the problem of data shortage can be partly solved by unsupervised learning methods with only sparse or even no annotated data, the ground-truth are still needed in experiments for evaluating or testing the generalization performance of the model. 
Thus, it is still of great significance to obtain a sufficient amount of images with accurate and dense depth information.
The common data acquisition method in real world is not feasible for the depth estimation, especially for aircraft visual navigation because humans cannot manually label a pixel-wise annotation.
Building a virtual world to generate synthetic datasets as the intermediate domain with the help of digital simulation technology may be the most feasible way for data generation and labeling at current stage.
Since the newly released synthetic datasets\cite{mayer2015iccv,sintel,blackbird,zhang2017physically} are not flexible enough to suit for different needs, e.g, fixed resolution, limited scenes, low data diversity and huge volume, etc, it is difficult to apply them to the dense estimation task in the large scale enviroment, especially for the depth estimation in aircraft navigation.
%
\begin{figure}[t!]
    \begin{center}
        \includegraphics[width=0.7\linewidth]{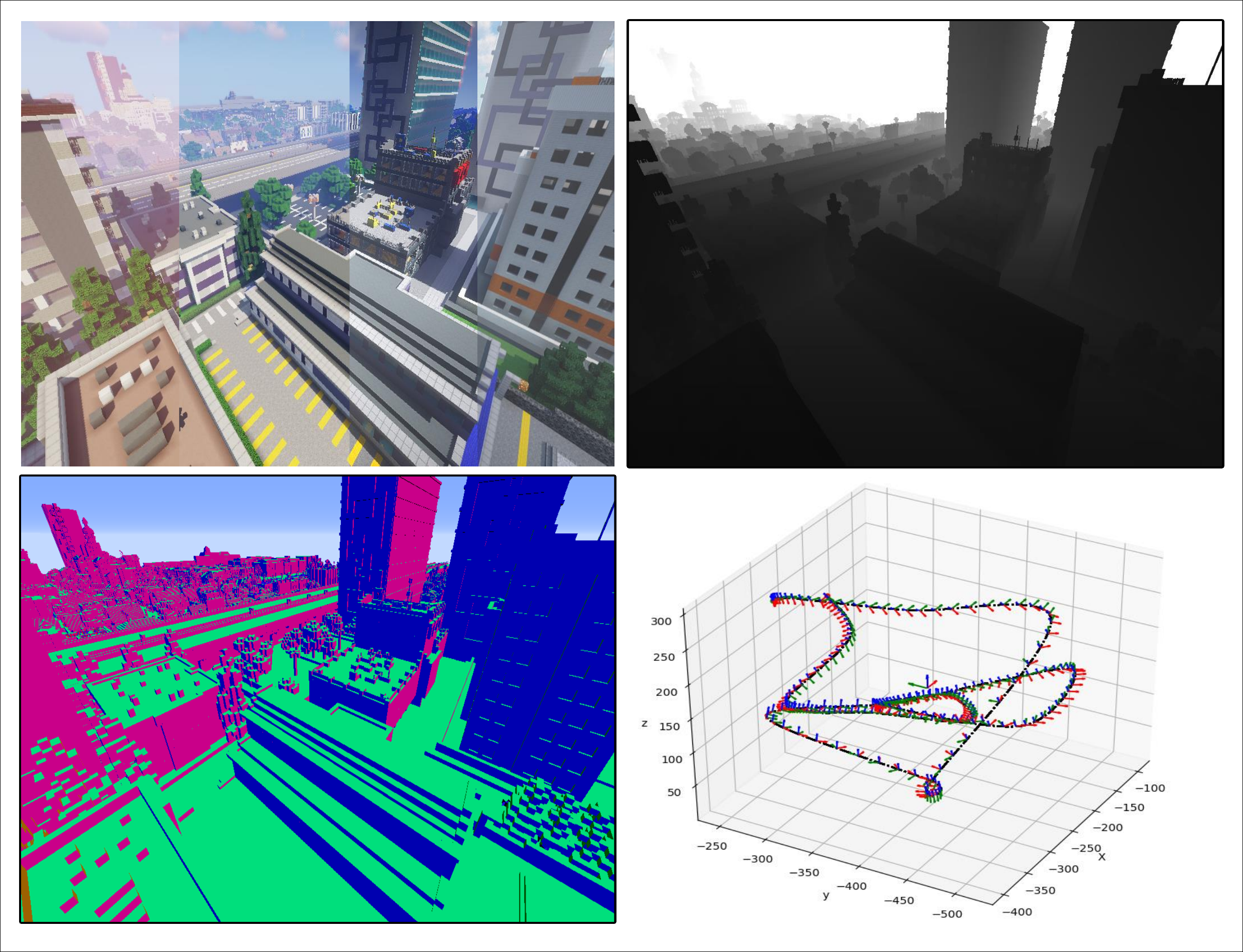}
    \end{center}
    \caption{MineNavi dataset provides image sequence, depth map, surface normal map and camera 6DoF pose in the large scale scene (with over 576m depth).}
        \label{fig:first}
  \end{figure}
Therefore, in this paper, we propose a simple and expandable synthetic dataset generation method, and construct a custom dataset, which is called as MineNavi (Figure \ref{fig:first}).
This dataset generation method can not only solve the problem of high cost of real-world data acquisition, but also can narrow the gap between the training domain and the target domain by customizing the synthetic scene that similar with the target domain.
Besides, different with conventional studies that adjust models in a fixed dataset to make them close to or superior to the state-of-the-art methods under certain evaluation metrics, we analyze the influences of the changes in datasets on the models. It is very significant because it can not only verify the generalization capabilities of the models to the environment, but also give guidance to construct real-world datasets.
In addition, to explore the impact of the various dataset factors on depth estimation models, our constructed MineNavi dataset contains the dense depth maps and surface normal vectors of objects. It will help us to observe the performance of depth estimation model under different factors of the dataset, such as the ego-motion camera, lighting and motion patterns, etc. Our experiments show that these variations on training sets may significantly affect the performance of the models.

Our contributions are as follows:
firstly, we propose an open synthetic dataset generation method and construct MineNavi for the large-scale depth estimation applications. 
Secondly, we design experiments to report the performance of the baseline models pre-trained on the MineNavi dataset, and reveal the influence of various factors in datasets on depth estimation models.
The source code is available.\footnote{\url{https://github.com/xdr940/MineNavi}}

\section{Related Work}
\label{sec:related_work}

\subsection{Synthetic Datasets}

Using synthetic data has a long history of application in computer vision community
\cite{ros2016synthia,vkitti,tremblay2018training,guerra2019flightgoggles,antonini2018blackbird}, 
it also is a significant data augmentation strategy for data driven models.
 On one hand, synthesis datasets solve the privacy issues involved in the data collection process partially.
 On the anther hand, the low-level dense prediction tasks, such as optical flow or depth estimation\cite{mayer2015iccv,sintel}, manual annotation is not feasible or even impossible, and synthetic datasets may be the only choice.
It is a better idea to use existing scenes from an open-source 3D movie to render artificial annotated data  for visual task\cite{sintel,mayer2015iccv}.
Using existing game scenes as data source is an ideal alternative\cite{wang2019learning,richter2016playing,guss2019minerl,blackbird}, which decrease the cost of manual 3D modeling of scene and realize capturing the data in arbitrary perspectives. 
Building datasets through open source games can openly and freely allow users to perform differentiated processing for specific vision task and generate the datasets they need, and through rich plug-ins, the cost and difficulty of building data sets can be further reduced to a certain extent.

\subsection{Bridging the Reality Gap}

Models trained purely on synthetic data often suffer limited generalization caused by domain gap between two type of datasets. Therefore, the utilization and theoretical analysis of the synthetic data set are also very necessary\cite{tremblay2018training}.
Domain Randomization (DR)\cite{tobin2017domain} is one of the most promising approaches to make straight-forward transfer learning from synthetic data to real world data.
 \cite{prakash2019structured} introduced structured domain randomization(SDR) to detection task, which imposes structure onto domain randomization (DR) in order to provide context, they have also shown that pre-training on SDR improves results from real data. 
\cite{metasim} represents the composition of a 3D scene with a scene graph and a probabilistic scene grammar, a common representation in computer graphics. 
Present works\cite{wang2018deep} have also devoted significant study to domain adaptation(DA), e.g., the problem of adapting models trained on source domain to a previously unseen target domain. 
In our proposed MineNavi dataset, we implement DR through multiple lighting conditions rendering, switch shader in the same scene and scene replacement of the same path etc.
Due to there is a little overlap between ImageNet and the dataset applied in large-scale depth estimation, MineNavi dataset can be ruled as a intermediate domain for multi-step domain adaptation.

\subsection{Datasets Strategy on MDE Training}
Although there is a domain gap between synthetic data and real data, most work is based on an assumption that the model trained on real data will perform some results that can be reproduced by training on synthetic data, which leads to the use of synthetic data as a proxy to explore the influence of factors in real data on the model.
\cite{vkitti} analyzed the multi-target detection task on different data with different illumination caused by weather conditions.
\cite{su2015render} analyzed the influence of background texture and lighting conditions on the task of detecting key points of the object, and concludes that more realistic lighting conditions are beneficial to the performance of the neural network in the task.
\cite{mayer2018ijcv} detailed analysis of the virtual dataset in the optical flow estimation task not only on the impact of lighting, but also included the shape movement and texture of the object.
While analysis on above datasets are fairly sufficient, what limits their performance is the lack of the the diversity among the benchmark datasets.
To this end, \cite{hu2020seasondepth} makes a synthetic dataset based on CMU Visual Localization dataset\cite{badino2011visual} and conducts extensive experimental evaluation on the proposed dataset with several learning-based algorithms. 
\cite{lasinger2019towards} introduced a strategy that leverage the mixed datasets im multiple scene for MDE's training.
Collecting Large-scale datasets from the Internet\cite{li2019learning} is also a way to increase the data diversity, but it require a huge manual labor for data pre- and post-processing.

In this article, we analyze the depth estimation task of weather and lighting conditions, and focus on the influence of camera self-motion, texture and imaging quality on unsupervised monocular depth estimation.
To the best of our knowledge, some factors that controlling data source has not been explored before in this context.

\begin{figure*}[t]
  \begin{center}
      \includegraphics[width=0.85\linewidth]{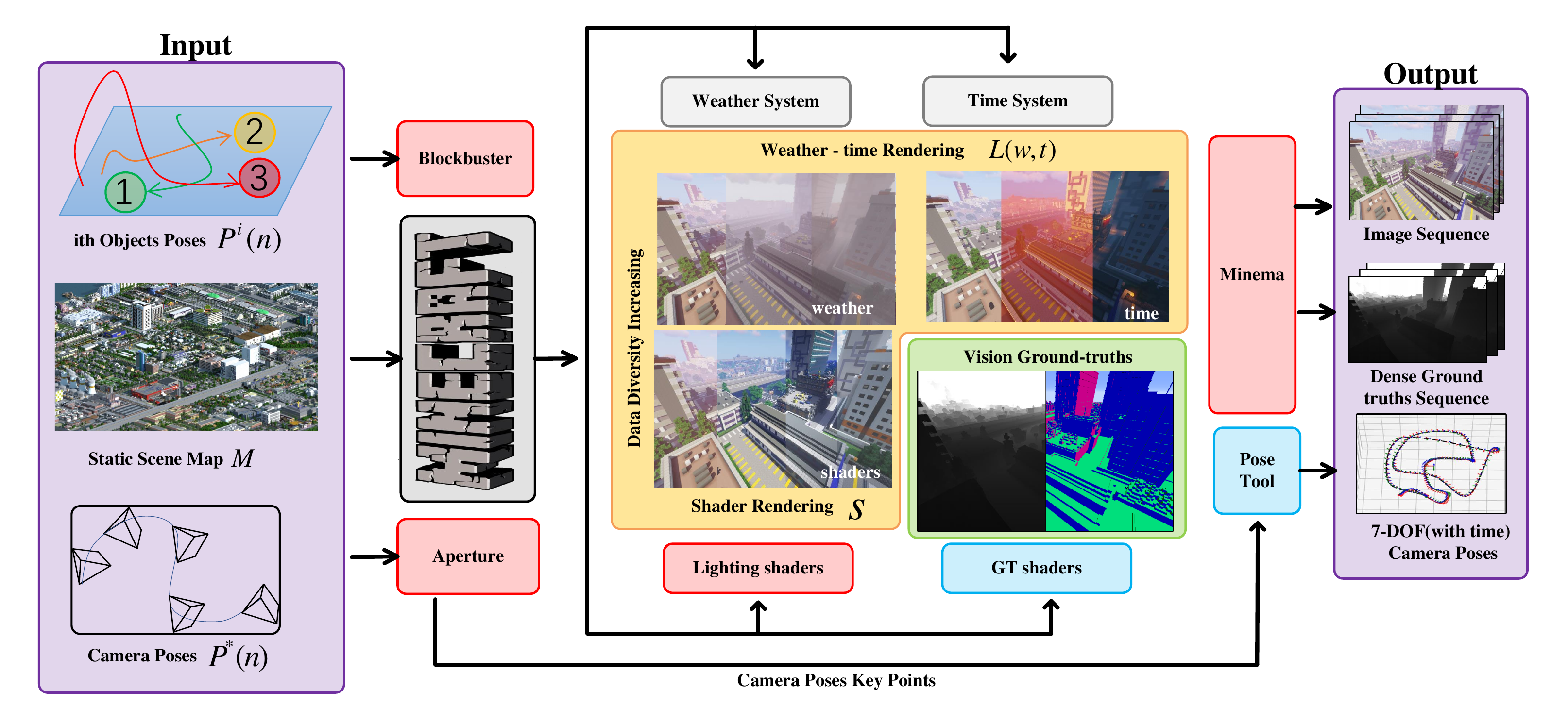}
  \end{center}
  \caption{Data generation pipeline. We use some open source tools and tools provided by ourselves to achieve efficient data generation.} 
      \label{fig:pipline}
\end{figure*}

\section{MineNavi: A Synthetic Dataset of Large Scale Scenes}
\label{sec:dataset}

Using MineCraft to construct a dataset is not a novel idea for computer vision community\cite{guss2019minerl}, we here to utilize it for depth estimation in aircraft visual navigation application. Our data generation method contains four steps: map loading, camera moving path setting, shader and lighting conditions setting, and ground-truth acquisition. \fig\ref{fig:pipline} shows the pipeline of data generation process.

Not only the environment features, such as the scene structure and lighting condition, affect the performance of depth estimation models, but also the particular dynamical parameters, such as moving targets in the environment and ego-motion of the aircraft, play important roles in benchmark datasets for models' training and evaluation.
Accordingly, each image frame of the dataset can be parameterized as:
\begin{eqnarray}
  I \left[ M,P^{\star}(n), s,L(t,w) \right]
\end{eqnarray}
  where $P^{\star}(n)$ represents the 6 Dof camera motion paths, $n$ is the quantified timestamp of the path, $M$ is the map of the scene, $s$ is the shader that renders the world, $L(t,w)$ is the lighting condition.
  $t$ is the time in a day, and $w$ indicates the weather conditions.
\subsection{Scene Construction}

Although a lot of work\cite{sintel,vkitti} build the scenes based on 3D modeling software such as \textit{Blender} and \textit{Maya}, the construction of large-scale 3D scenes is still a relatively time-consuming and laborious task.
Besides, the limited scenes diversity will lead to the over-fitting situation of the models. 
\textit{MineCraft} community\cite{minecraftmaps} has extremely rich scene maps and users can freely build the required scene to generate specific dataset. Since the aircraft navigation is always involved in the large-scale scenes, the negative effects of the jagged features of objects in MineCraft can be ignored. 

In order to increase the diversity of data, we use different shaders and lighting conditions. MineNavi dataset cooperates with the time and weather system according to different light and shadow styles to generate multiple style data.

\begin{figure}[t!]
  \begin{center}
      \includegraphics[width=.95 \linewidth]{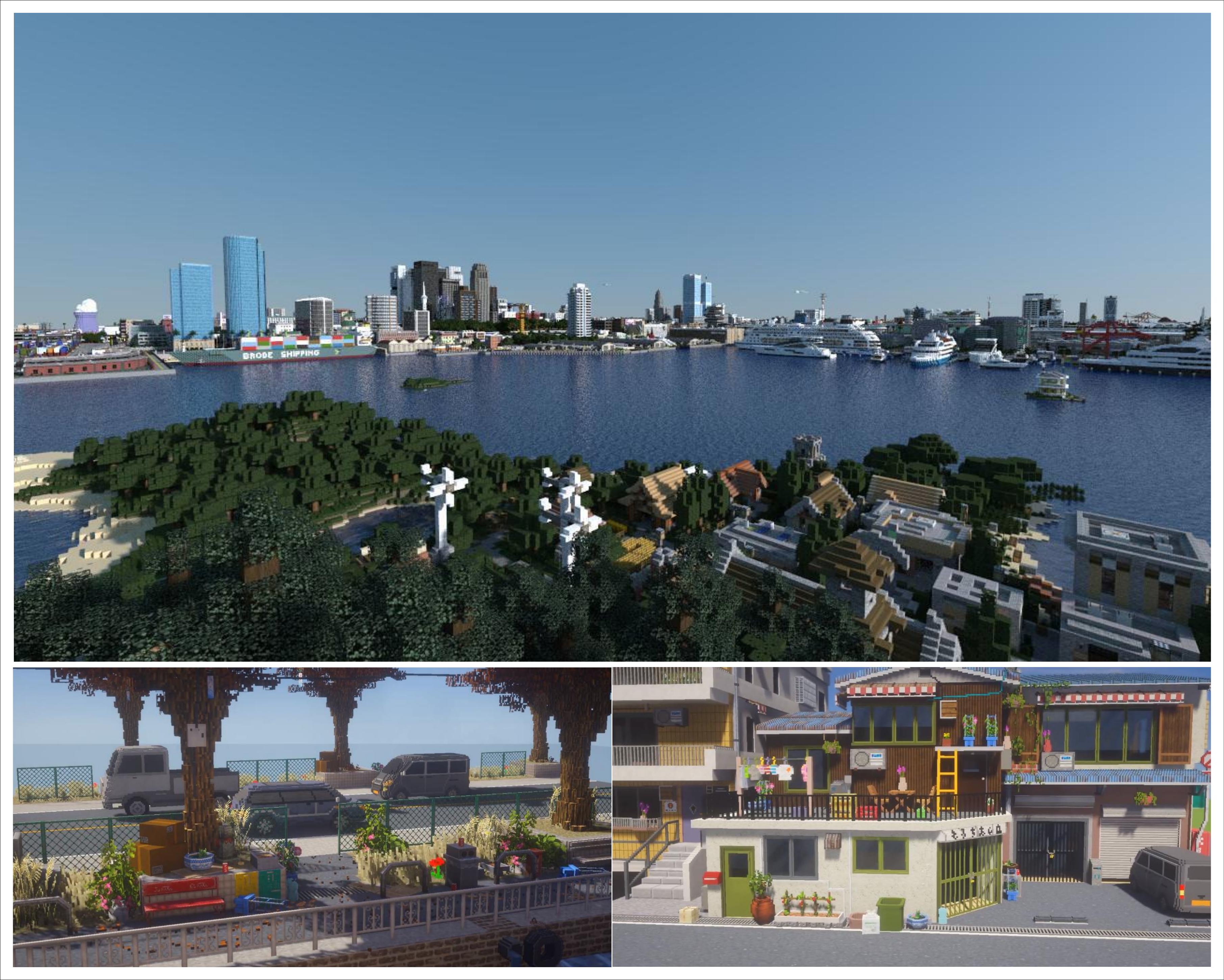}
  \end{center}
  \caption{Virtual world constructed in \textit{MineCraft}. Up: The open virtual world \textit{AudiaCity} that we used to build our dataset. Down: Users can achieve higher resolution scenes or buildings by applying plugins that adjust the blocks to small size.} 
      \label{fig:audia}
\end{figure}

The construction in MineNavi based on the block is very simple and flexible. In order to build a more refined scene, users can use plug-ins to adjust the size of the block to achieve more complex objects (See \fig\ref{fig:audia}).

\subsection{Camera Paths Setting}
  \label{sec:camera-motion}
  Base on previous study\cite{bian2019unsupervised,zou2018df,packnet}, we have found that the unsupervised monocular depth estimation methods are very sensitive to the camera motion in the training.
  Therefore, we develop different camera paths and generates corresponding datasets for experiments.
  Unlike lighting and other factors that can be quantified as a scalar, a moving camera has 6 continuous degrees of freedom.
  Therefore, for a training triplets, we propose a quantitative scalar $\lambda$, i.e., quasi-axis rate to generate datasets of the motion paths according to $\lambda$, and analyze the pros and cons of the data under different $\lambda$. The $\lambda$ can be formulated as:
  \begin{eqnarray}
      \lambda(n) & = &\frac{\bar{\phi}(n) \dot{t}(n) }{\| \bar{\phi}(n)\|\|  \dot{t}(n)\|}\\
      \bar{\phi}(n) &= &\frac{\phi(n+1) + \phi(t-1)}{2} \\
      \dot{t}(n)&= &t(n+1) - t(n-1)\\
      P^\star(n)& =& (\phi(n),t(n) )
  \end{eqnarray}
  $\phi(n) $ is the rotation angle of camera visual axis  at time $n$, calculated from $R\in SO(3)$,and $t(n) = [x,y,z]$ is the camera position vector at time n.
   When $\lambda =0$, the camera moves parallel to the visual axis, when $\lambda =1$, the camera moves perpendicular to the visual axis.

     
  \begin{figure}[t!]
      \begin{center}
          \includegraphics[width=1.0\linewidth]{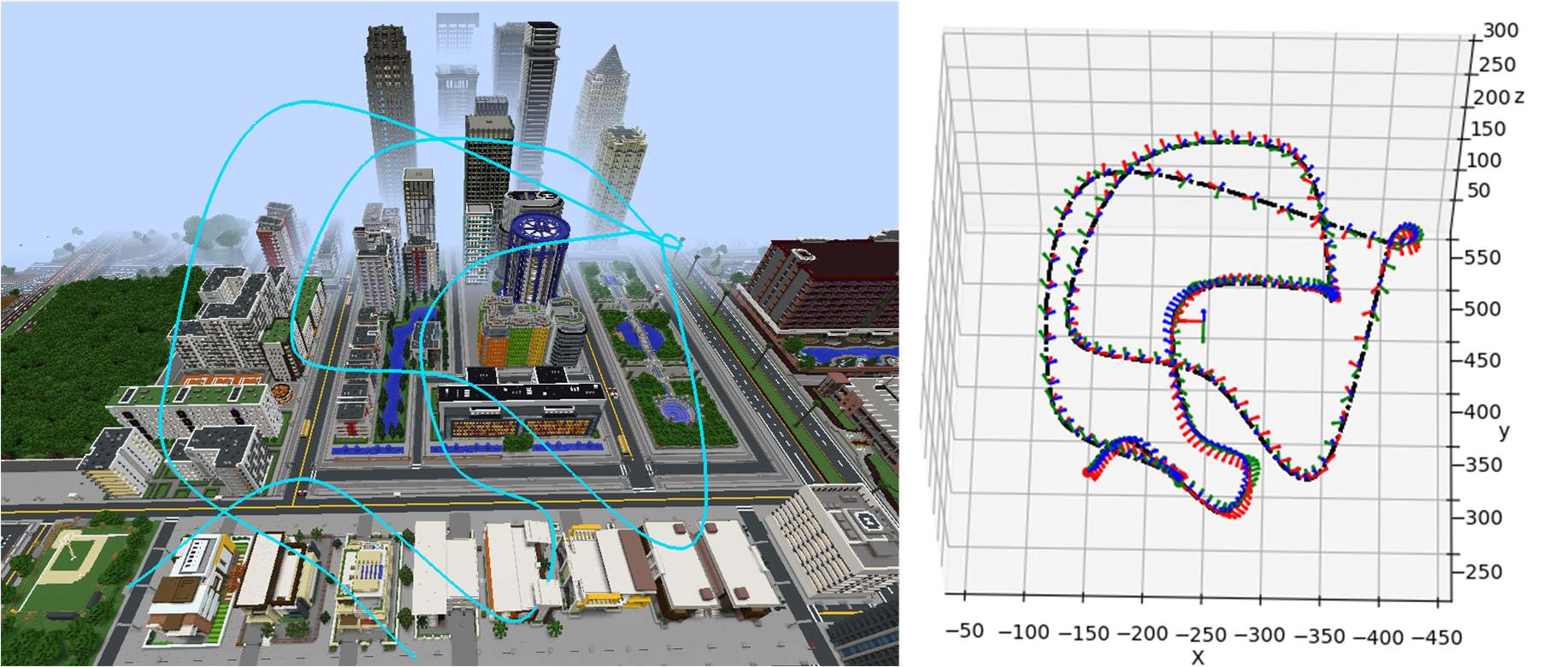}
      \end{center}
      \caption{ The accurate camera 6Dof pose at any timestamp in the path is gotten by exporting the key points of the \textit{Aperture} path and interpolating.}
          \label{fig:traj}
    \end{figure}
In MineNavi, we can set the key points manually or automatically by using \textit{Aperture}\footnote{ \url{https://www.curseforge.com/minecraft/mc-mods/aperture}} and obtain the full path that matched with image sequence through the interpolation algorithm (See \fig\ref{fig:traj}). The generated path has high enough dynamics, and the pose transformation is much larger than the general real-world data captured by UAVs.


\subsection{Moving Objects in the Scene}
The dynamic objects in a practical environment may have a great influence on unsupervised monocular depth estimation method. 
Many previous work\cite{yin2018geonet,watson2021temporal,klingner2020self} focused on how to remove the negative effects of dynamic objects in monocular depth estimation, but due to the very limited dataset, the progress is far from satisfactory. 
In order to simulate the influence of a moving object in the synthetic dataset on the depth estimation, the construction of the scene containing moving objects can be further parameterized as:

\begin{eqnarray}
  M\left[ P^1(n),P^2(n),\ldots,P^m(n)  \right] 
\end{eqnarray}
where $P^i$ is the path of the ith moving objects in the scene.
Each of the dynamic object can be modeled as custom shape by \textit{Blender} or other 3d software and set their paths by using \textit{BlockBuster}\footnote{https://www.curseforge.com/minecraft/mc-mods/blockbuster}. Note that we have no involved the moving object in our proposed dataset yet due to its negative effects on depth estimation model, more details are given in the supplementary material.



\subsection{Generating ground-truth annotations}
  
 The shader can perform color mapping on the 3D information of the scene, which acts as a ground-truth as shown in \fig\ref{fig:first}.
  We use the \textit{DepthMap} rendering plug-in to export the corresponding error-free, pixel-level dense depth map that matches the image in sequences. In addition, we provide a surface normal rendering plug-in \textit{SurfMap} to support surface normal estimation tasks.

  
  Thus, the datasets construction method proposed in this study can generate a large number of customized datasets at a very low cost. Related code and details are shown in supplementary material.

\section{Preliminary}

In this section, we mainly talk about the principle of MDE models, a analysis method of ego-motion video for MDEs and the relevant dataset for the training and evaluation of MDEs in our experiments.

\subsection{Deep Models for Monocular Depth Estimation}

        In order to reconstruct 3D geometry from the ego-motion video by self-supervised method, the main supervision signal comes from the view synthesis between consecutive frames.
         We aim to infer the intermediate variable, e.g. predicted depth as precisely as possible through this approach.
        This is an ill-posed problem because a single 2D image may be produced from a large number of distinct 3D scenes, thus it can only be resolved using prior knowledge about the appearance and motion of image sequences.

         Assume that $I_t$ and $I_{s}(I_{s} \in \{I_{t-1},I_{t+1} \} )$ are two consecutive frames from an unlabeled video, we want to establish the dense pixel correspondence between the two frames. 
        Let $p_t$ denotes the 2D homogeneous coordinate of a pixel in frame $I_t$ and $K $denotes the intrinsic camera matrix. We can compute the corresponding point of $p_t$ in frame$I_s$ using the following equation,
        \begin{eqnarray}
        \quad &p_{s} = K\hat{T}_{t \rightarrow s}\hat{D}_{p_t}K^{-1}p_t
        \end{eqnarray}
        where $\hat{D}_t$ is the predicted depth map, and $T_{t\rightarrow s}$ is the relative camera pose. We can get the synthesized image $\hat{I}_s$, $I_s = I_t(p_s)$ Thus, the photometric error can be formulated as:
        \begin{eqnarray}
        \quad &\mathcal{L}_p = \sum\limits_{s \in \{t^-, t^+\}}\sum\limits_p\rho(I_s(p), \hat{I}_s(p))&
        \end{eqnarray}
        where $ \rho(\cdot)$ is a function to measure the difference of pixels value in the two images. We use a combination of \textit{SSIM}\cite{wang2004image} and L1 term as our photometric error function which is formulated as:
        \begin{eqnarray}
            \rho(I,\hat{I}) = \alpha \frac{1-\mathcal{SSIM}(I,\hat{I})}{2} + (1-\alpha)\|I-\hat{I}\|
        \end{eqnarray}

        It is robust to illumination changes in a real-world scenario and we set $\alpha = 0.85$. In general, the photometric loss $\mathcal{L}_p = \rho(I,\hat{I})$.

        However, the supervision signal just based on view synthesis is not informative enough in low-texture or homogeneous regions of the scene due to its ambiguities. Thus, additional regularization is required to learn reasonable depth prediction. A common strategy proposed by \cite{sfmlearner,monodepth}, and it is formulated as :
        \begin{eqnarray}
        \quad&L_{s} = \sum\limits_{p_t}\sum\limits_{d \in x,y} \| \nabla_d^2 \hat{D}(p_t) \|_1 e^{-\alpha|\nabla_d I_t(p_t) |}&
        \end{eqnarray}
        where $\nabla$ is the first derivative along the spatial direction. $L_s$ ensures that the edges in depth map are guided by the images.And overall objective function can be formulated as follows:
        \begin{eqnarray}
        \quad&\mathcal{L}^l_{final} = \mathcal{L}^l_p ~+~\lambda\mathcal{L}_{s}^l&
        \end{eqnarray}
        where $l$ stands for multi-scale approach.

\begin{figure*}[t!]
    \begin{center}
        \includegraphics[width=0.95\linewidth]{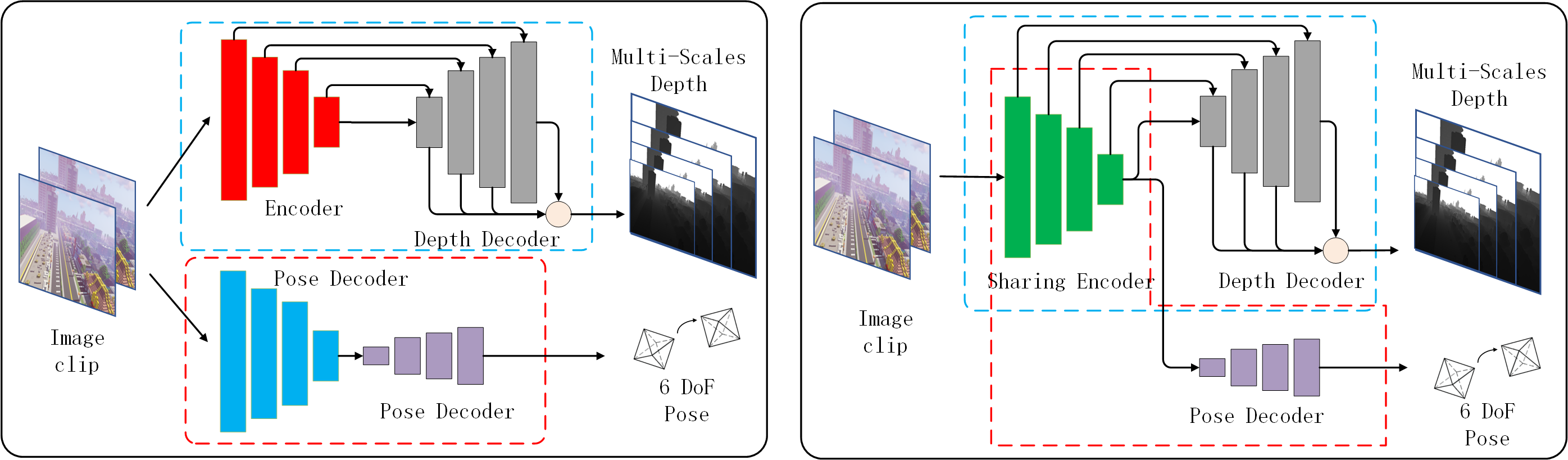}
    \end{center}
    \caption{For spatiotemporal feature learning, We build a 3d encoder and apply it into monodepth2 to build monodepth2-3D (Left) and monodepth2-3Ds (Right)}

        \label{fig:mtl}
  \end{figure*}

Unsupervised depth estimation includes monocular methods\cite{sfmlearner,monodepth,packnet,zou2018df,bian2019unsupervised,cc,yang2018lego} usually contain a single-view depth and a multi-views pose network, to compute the depth. With the similar principle, we use a test model monodepth2\cite{monodepth2} and its variants as baseline.


Inspired by spacial-temporal methods in scene understanding\cite{li2020spatial,li2020spatio},
the first variant of monodepth2 is monodepth2-3D, i.e., replace the encoder with a 3D encoder for improving the efficiency of training frames, which can enhance the richness by extracting the temporal features from multiple images\cite{watson2021temporal,hur2021self,patil2020don}.
What's more, as mentioned by previous work\cite{taskonomy} that if there is structural similarity among candidate tasks, it is reasonable to assign just one encoder to extract identical features and recover required information by task-oriented decoders respectively\cite{sfmlearner,monodepth2}.
Thus, we apply the model that using a single encoder to extract the mixed features for depth or pose estimation network as second variants of monodepth2, i.e., monodepth2-3Ds.

\subsection{The Sequential Heat-map of Photometric-error Histogram }

\begin{figure}[htp]
    \begin{center}
        \includegraphics[width=0.95\linewidth]{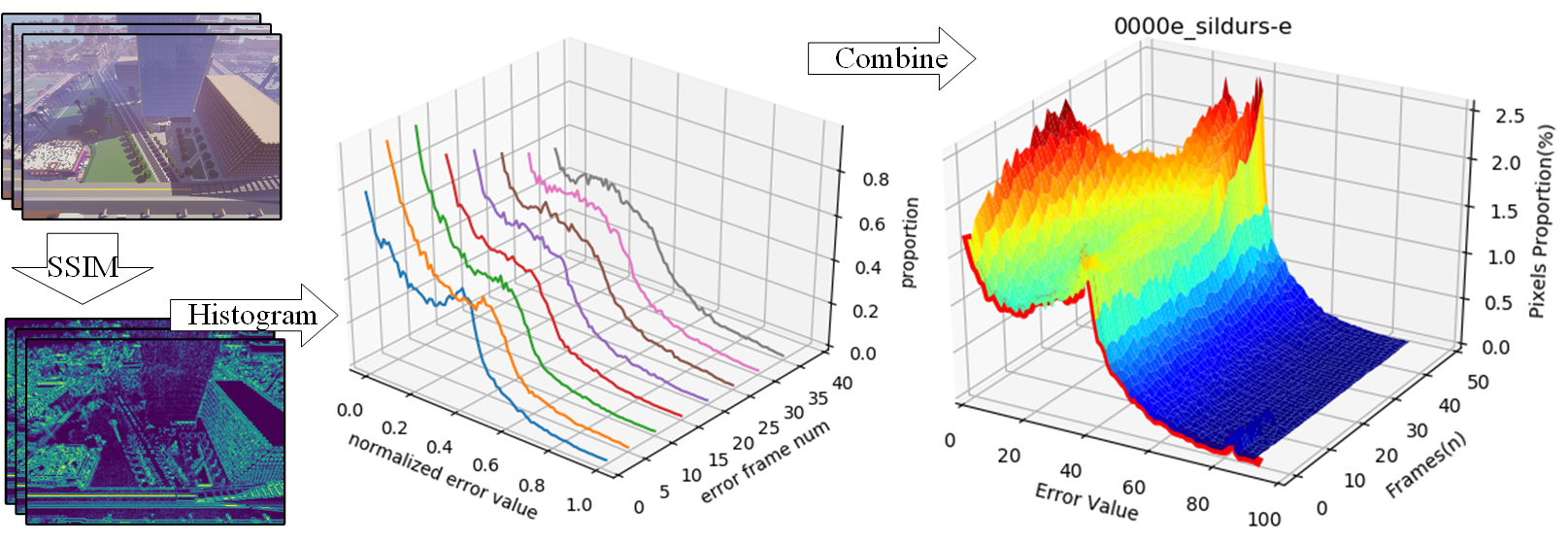}
    \end{center}
    \caption{The process of SHPH. In an image sequence, by normalizing the histogram statistics of any adjacent frame photometric error map, and then connecting together, the heat map that reflect the suitableness of training set to MDE could be obtained.}

        \label{fig:sheh}
  \end{figure}

  The unsupervised monocular depth estimation model reconstructs the depth of the scene based on the principle of view synthesis\cite{garg2016unsupervised,flynn2016deepstereo}, i.e. sfmlearner\cite{sfmlearner}, which offer the main supervisory signal during the training.  
  The quality of the view synthesis seriously affects the performance of the model. Thus, we propose a Sequential Heat-map of Photometric-error Histogram (SHPH) to visualize a image sequence (See \fig\ref{fig:sheh}) and verify weather it is compatible with depth estimation model training intuitively.
  
  We have found that during the training, the depth estimation models are very sensitive to the distribution of SHPH, The reason behind is that if the photometric error value in the images is evenly distributed, the model can perform more stable during gradient back-propagation, thereby improving the training efficiency.
  Accordingly, the histogram of each photometric error map in the training sequence should be kept as even as possible, which also provides a basis for choosing a suitable training sequence.

\subsection{Datasets}
 The depth estimation method that based on SfM is not suitable for some datasets due to the absent of successive image\cite{li2018megadepth,Silberman:ECCV12}, large scene scale\cite{sturm12iros} or applicable camera motion\cite{zhu2018vision,barekatain2017okutama,euroc}.
 Therefore in our experiments, we use the follow datasets to make comparison and evaluate the feasibility of the dataset generation method.
\begin{itemize}
    \item \textbf{KITTI}  is widely used in monocular depth learning. In this study, we use KITTI Zhou split\cite{sfmlearner} that contains image triplets, which static frames with an average optical flow of less than 1 pixel are removed. 
    In our experiment 40k triplets are used for training and 4k triplets for testing. 
    \item \textbf{FPV.} To test the performance of depth estimation models in the large-sale scenes, we selected the video data of the first person view (FPV) of the aircraft as another experimental dataset.
    These dataset cover most of the field scenes, including mountains, woods, lakes, snow mountains and cities. 
  Different with previous uav datasets\cite{barekatain2017okutama,zhu2018vision} which captured by simple or even static camera motion, this dataset has diverse moving pattern that suitable for depth estimation model based on SfM.
    We divide the data into two datasets with the train set volume is 36k and 4k validation set.
    \item \textbf{MineNavi} datasets contain data samples according to different shaders(middle-sildurs, high-sildurs), lighting conditions (morning, moon, evening, night, sunny, and rainy), camera motions(linear motion and circular motion) and image qualities (clear, low blur, middle blur, high blur). We set the field of views of camera as 70 degree. The resolutions of the captured images are $1024 \times 768$. The matched depth map ranges from $0.1m$ to $576m$.
    Finally, we set more than 400 paths around a virtual scene of more than 16 square kilometers, and collected a total of more than 50,000 image sequences. The image scenes include various weather and lighting conditions, with matching depth values.
\end{itemize}
\begin{figure}[h]
    \begin{center}
        \includegraphics[width=0.95\linewidth]{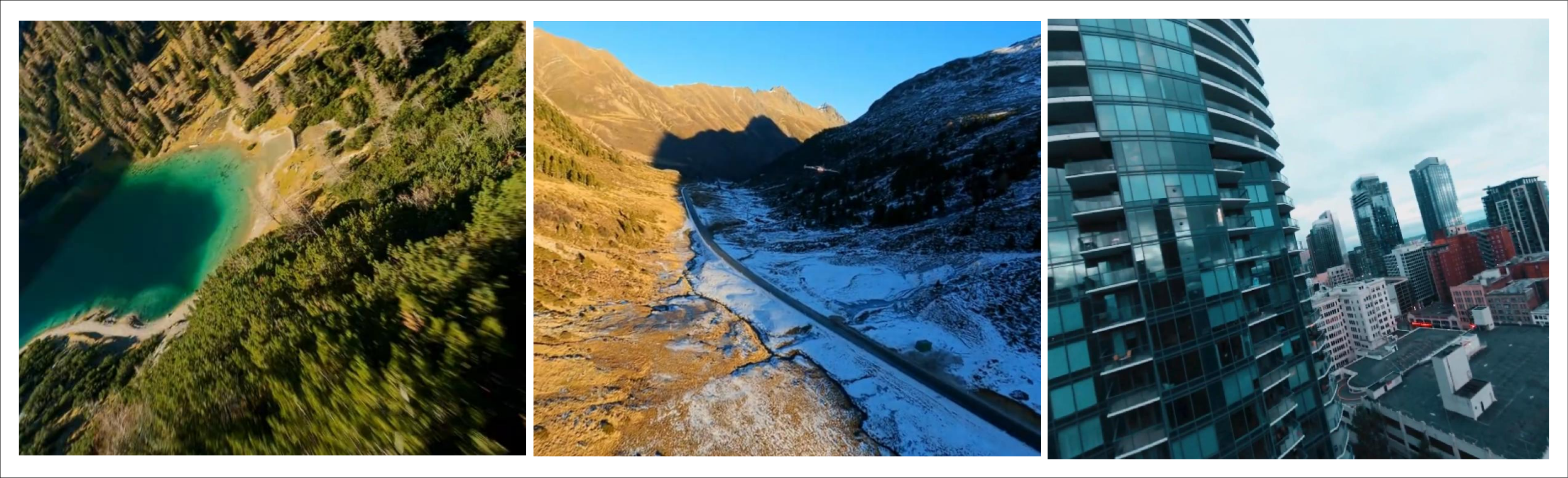}
    \end{center}
    \caption{FPV datasets. Left two: Fpv-filed. Right: Fpv-downtown. We introduce FPV dataset to test the generalization of proposed  MineNavi dataset.}
        \label{fig:fpv}
\end{figure}

\subsection{Evaluation Metrics}
For depth estimation, we evaluate depth estimation models using three accuracy metrics ($\delta^1,~\delta^2,~\delta^3$) and four error metrics : absolute relative difference (AbsRel), square related difference (SeqRel), RMSE, log RMSE. 
Note that the accuracy metrics and $AbsRel$ are scale-independent, i.e., the value does not change with the depth range of the dataset. And the others are scale-invariants, i.e., the results evaluated on the dataset with  different depth ranges are in different order of magnitudes.

\section{Experiments}
\label{sec:exp}

In this section, we verify the feasibility and credibility of the MineNavi dataset in the training of the depth estimation model, and explore the impact of dataset varieties on the unsupervised monocular depths estimation model. Thus, we will demonstrate that
1) the depth estimation model can improve generalization through pre-training on MineNavi.
2) it is desirable to exploit the influence of data to model caused by various factors of the dataset.
We prepare monodepth2\cite{monodepth2} and its two variants monodepth2-3D and monodepth2-3Ds as the test models on our proposed datasets. We also present using Sequential Heat-map of Photometric-error Histogram (SHPH) to verify weather an image sequence is compatible with depth estimation model training intuitively.
The detailed description is given in the supplementary material.



\subsection{Apply MineNavi to MDE Models}
\begin{figure*}[t!]
    \begin{center}
        \includegraphics[width=1.0\linewidth]{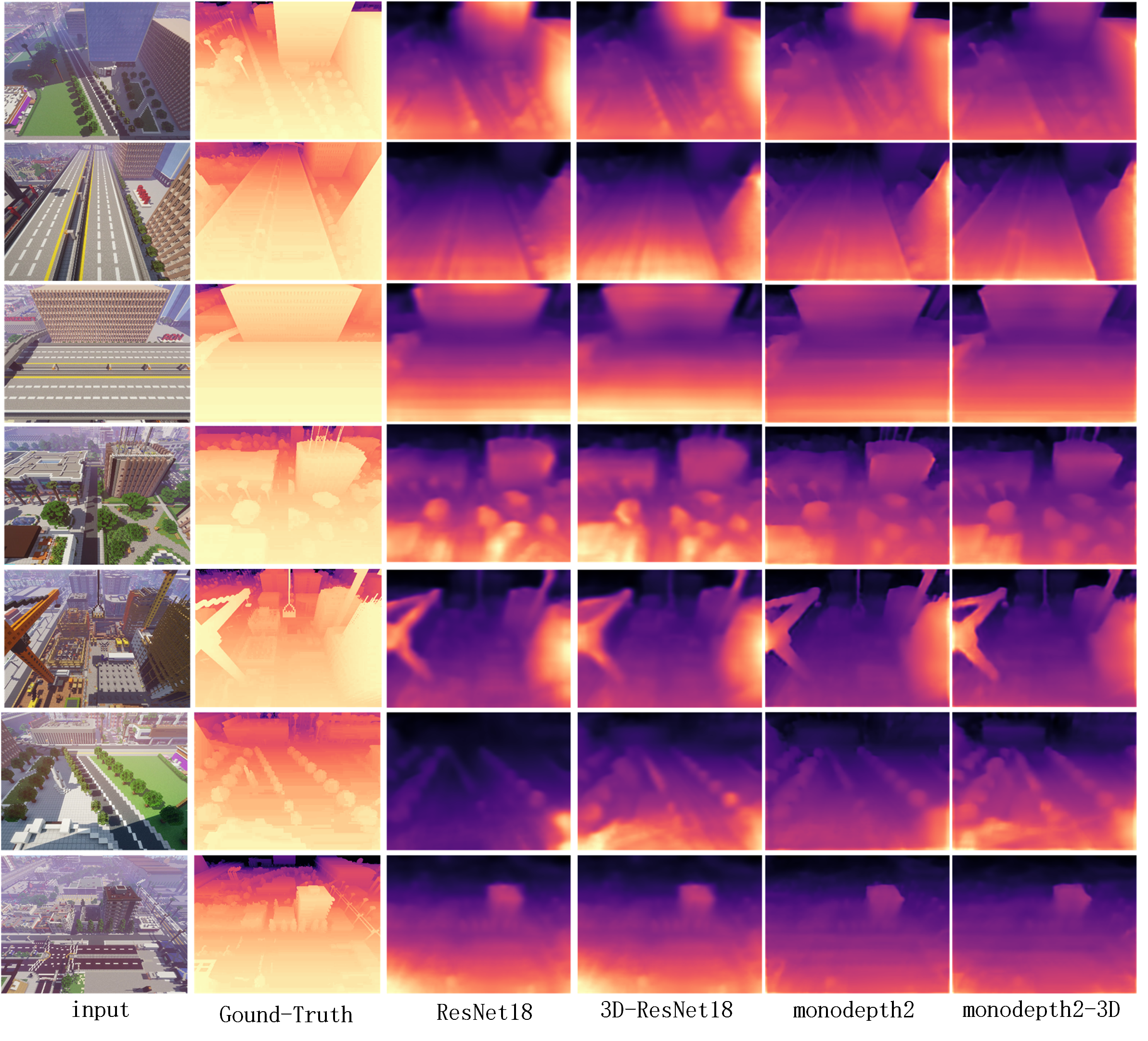}
    \end{center}
    \caption{Qualitative Results of different models in MineNavi.}

        \label{fig:qualitative-models}
  \end{figure*}

We present two variant models by changing their encoder, and apply them into frameworks of supervised training and unsupervised training (monodepth2).
\tab\ref{tab:exp-encoders} shows the superiority of multi-frame input models: by simply replacing the encoder from ResNet18 to 3D-ResNet18 and using supervised training method (the model components are identical with monodepth2), they all achieved similar or even better results, and the amount of parameters did not increase, or even reduced by half (monodepth2-3Ds).
The quantitative results are shown in \fig\ref{fig:qualitative-models}

\begin{table*}[htp]
    \caption{Quantitative results in MN datasets. First six model are trained in supervised and the rest are unsupervised (monodepth2). } 
    \label{tab:exp-encoders}
    \centering
    \scalebox{1.0}{
    \begin{tabular}{ll|cccc|ccc}
       \toprule[2pt] 
      && \multicolumn{4}{c|}{Error $\downarrow$} & \multicolumn{3}{c}{Accuracy $\uparrow$}  \\
      \cline{3-9}
      Model& dataset & AbsRel & SqRel & RMS & RMSlog & $<1.25$ & $<1.25^2$ & $<1.25^3$ \\\hline

      ResNet18 
      &MN-2K(middle)&0.198&1.318&5.679&0.268&0.731&0.923&0.972\\
      3D-ResNet18
      &MN-2K(middle)& \textbf{0.194} &\textbf{ 1.197} & \textbf{5.453} &\textbf{ 0.259} & \textbf{0.749} & \textbf{0.932} & \textbf{0.973} \\

      ResNet18 
      &MN-2K(high)&0.207&1.479&5.832&0.274&0.721&0.920&0.968\\
      3D-ResNet18
      &MN-2K(high)&\textbf{ 0.181 }&\textbf{ 1.159} & \textbf{5.372} & \textbf{0.250} &\textbf{ 0.761} &\textbf{ 0.935} & \textbf{0.976} \\

      ResNet18 
      &MN-10K(high)&0.142&0.707&4.455&0.181&0.846&0.961&0.986\\
      3D-ResNet18
      &MN-10K(high)& \textbf{0.129} &\textbf{ 0.675} & \textbf{4.384} &\textbf{ 0.172} & \textbf{0.861} & \textbf{0.966} & \textbf{0.987} \\
      \toprule[0.5pt]
      
     monodepth2 
      &MN-2K(high)&\textbf{0.212}&\textbf{1.426}&7.054&0.295&0.706&0.902&0.959\\
      monodepth2-3D
      &MN-2K(high)& 0.245 & 1.833 & \textbf{5.919} & \textbf{0.273} & \textbf{0.750} & \textbf{0.925} &\textbf{ 0.965} \\

      monodepth2
      &MN-10K(high)&0.170&\textbf{0.974}&\textbf{5.782}&0.211&0.798&0.941&0.977\\
     monodepth2-3D
      &MN-10K(high)& \textbf{0.165} & 1.170 & 5.965 &\textbf{0.211} & \textbf{0.800} & \textbf{0.942} & \textbf{0.977} \\

      monodepth2-3Ds
      &MN-10K(high) &\textbf{0.160}&0.991&5.899&\textbf{0.208}&\textbf{0.809}&\textbf{0.943}&\textbf{0.977}\\

      \toprule[2pt]
    \end{tabular}
    }
\end{table*}

It is completely inconsistent with the results of KITTI that multi-frame input models achieve worse results, we suspect that it is because the scenes in KITTI have more moving objects that contradict with Warping-based view synthesis principle, therefore, it may reveal that compared to image input, simple multi-frame input model is superior than single input models in the scenes if the problematic pixels are filtered better.

\subsection{Generalization of MineNavi}
\label{sec:exp:gnz}

We execute the models pre-training on MineNavi datasets with linear camera moving path. 
In order to evaluate the influence of data diversity on the performance, we collect two data samples:
the first one contains 10k samples (MN-10K) rendered by high-sildurs shader with the sunny weather, and the second one contains 50K samples (MN-50K) rendered by three shaders (None, middle-sildurs, and high-sildurs) with various lighting conditions (morning, moon, night, and sunny). For comparison, we also prepare the models that are trained from scratch and ImageNet\cite{ImageNet}. 
Although the model pre-trained on ImageNet by classification task has structural difference compare with the model that trained on similar target task, it is still the most popular method in depth estimation task.

\subsubsection{Fine-tune on KITTI}

We conduct fine-tuning of the monodepth2 and monodepth2-3Ds pre-trained from above datasets on KITTI dataset for 10 epochs.
\begin{figure*}[h]
    \begin{center}
        \includegraphics[width=1.0\linewidth]{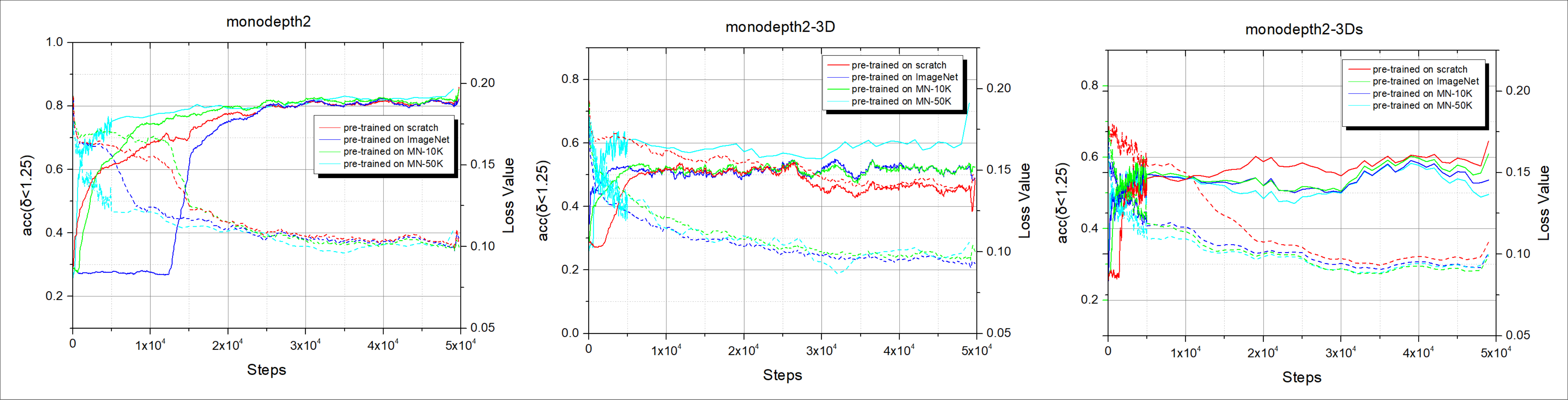}
    \end{center}
    \caption{Fine-tuning curves of three test models on KITTI. Solid curves denote accuracy ($\delta \leq 1.25$) metric of depth estimation and dash curves denote training loss.}
        \label{fig:kitti-gnz}
\end{figure*}
\begin{table*}[htp]
  \caption{Quantity Results in KITTI with different pre-trained datasets. The best result are bolded and the second best are underlined.} 
  \label{tab:kitti-gnz}
  \centering
  \scalebox{0.9}{
  \begin{tabular}{l|l|cccc|ccc}
     \toprule[2pt] 
    && \multicolumn{4}{c|}{Error $\downarrow$} & \multicolumn{3}{c}{Accuracy $\uparrow$}  \\
    \cline{3-9}
    Models& Pre-trained Datasets & AbsRel & SqRel & RMS & RMSlog & $<1.25$ & $<1.25^2$ & $<1.25^3$ \\\hline
    
    \multirow{4}{*}{monodepth2}
    &scratch &      0.141     & 1.117          & 4.797 & 0.205 & 0.839 & 0.948 & 0.980 \\	
    &ImageNet&      \underline{0.135 }    & \textbf{1.007} & \underline{4.668} & \underline{0.200} & \underline{0.845 }& \underline{0.950} & \textbf{0.980} \\
    &MN-10k	&       0.138     & 1.095          & 4.722 & 0.204 & 0.843 & 0.949 & 0.979 \\
    &MN-50k	& \textbf{0.130}  & \underline{1.055} & \textbf{4.630} & \textbf{0.196} & \textbf{0.856} & \textbf{0.953} &\textbf{ 0.980} \\
    
    \toprule[0.5pt]

    \multirow{4}{*}{monodepth2-3D}
    &scratch& 0.170 & \textbf{1.453} & 5.758 & 0.247 & 0.775 & 0.916 & 0.963 \\
    &ImageNet& \underline{0.163} & 1.857 & \underline{5.529 }& 0.233 & 0.807 & 0.930 & 0.968 \\
    &MN-10k& 0.167 & 2.468 & 5.671 & \underline{0.230} & \underline{0.820} & \underline{0.932} & 0.966\\
    &MN-50k& \textbf{0.153} &\underline{1.639} &\textbf{ 5.356 }&\textbf{ 0.224 }& \textbf{0.820} & \textbf{0.936} & \textbf{0.970}\\
    \toprule[0.5pt]

    \multirow{4}{*}{monodepth2-3Ds}
    &scratch& 0.161 & \textbf{1.660} &\textbf{ 5.449} & 0.228 & 0.805 & 0.930 & 0.970\\
    &ImageNet& \textbf{0.158} & 2.447 & \underline{5.511} & \textbf{0.220} & \textbf{0.839} & \textbf{0.938} &\textbf{ 0.970} \\
    &MN-10k& 0.165 & 2.361 & 5.535 & 0.229 & 0.820 & 0.934 & 0.968\\
    &MN-50k & \textbf{0.158} & \underline{2.133} & 5.555 & \underline{0.223} & \underline{0.829} & \underline{0.936} & 0.969\\

    \toprule[2pt]
  \end{tabular}
  }
\end{table*}

From \tab\ref{tab:kitti-gnz}  it can be seen that the performance of monodepth2 and monodepth2-3D pre-trained on ImageNet is better than that pre-trained on MN-10k and scratch, but worse than MN-50k. The MineNavi has a strong generalization capability compared to the KITTI. As mentioned before, MN-10k and MN-50k are only different in lighting condition and data volume. Therefore, the diversity of lighting conditions effectively improves the generalization capabilities of the models. 
Compared with the other datasets, the model of monodepth2-3Ds pre-trained on ImageNet has the better performance. This is mainly because excessive noises in KITTI, e.g, the moving objects deteriorate the robustness of the network performance of the shared encoder, but the large amount of data of ImageNet can make the model more robust\cite{hendrycks2019using}. Note that although MineNavi dataset is much smaller the ImageNet, it has competitive performance with ImageNet in depth estimation model training.
We also provide qualitative result on \fig\ref{fig:kitti-gnz} and it shows the value in generalization of our MineNavi dataset. 

\subsubsection{Fine-tune on FPV}

Since there is no ground-truth in the FPV dataset, we have to compute the distances between the models in different domains based on the loss value\cite{taskonomy}. The closer the migration distance is, the better the pre-training dataset can be generalized to the target domain.


\begin{figure*}[htp]
    \begin{center}
        \includegraphics[width=1.0\linewidth]{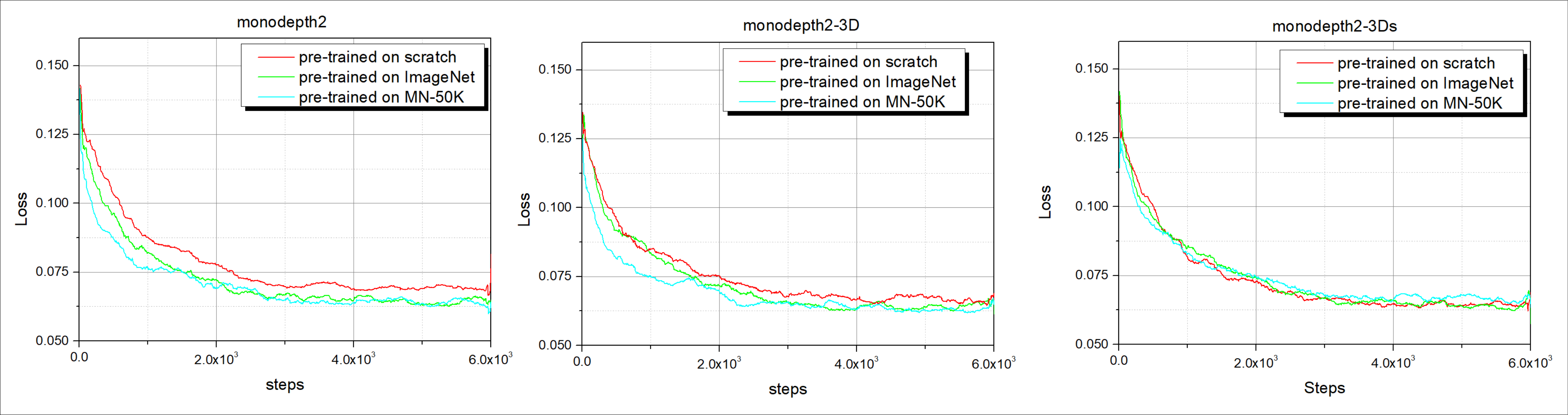}
    \end{center}
    \caption{Fine-tuning curves of three test models on FPV.}
        \label{fig:fpv-gnz}
\end{figure*}

 Compared the losses curve among of different pre-trained models that are fine-tuned on FPV in \fig\ref{fig:fpv-gnz}, MineNavi pre-trained models converge faster than the others. 
The reason behind probably comes from that the MineNavi dataset is closer to the FPV dataset than ImageNet in terms of environment scenes. 
What's more, compared with the ImageNet pre-trained model through the task of objects detection, the MineNavi pre-trained model through the task of depth estimation has learned geometric representation\cite{wang2020geometric} during the pre-training, which leads the model converge faster when the target task has structural similarity\cite{taskonomy} with source task.
 Note that, with the continuous expansion of the dataset, MineNavi can realize a more satisfied performance.


\subsection{Factors that Affect the Train of MDE}
\label{sec:factors}

Due to the expandable characteristics of the MineNavi dataset, we can easily generate customized datasets with different variation factors to avoid the over-fitting. 
It also a helpful wat to discover the impacts of factors of datasets on the models. 
 Thus we conduct experiments to explore how the factors in dataset can affect MDE model, including the shader, lighting conditions, motion blur, ego-motion and velocity of training image sequence.

\subsubsection{Impact of Shaders}

The MineNavi dataset can generate the rendered image sequences sampled on the same path through different shaders, which allows us to quantitatively evaluate the impacts of the synthetic world design and the quality of other rendering parameters on the algorithm performance. 
We apply ~Sildurs~\footnote{https://sildurs-shaders.github.io/} to adjust the image rendering quality and build three datasets \textit{Raw, middle-sildurs} and \textit{high-sildurs}.
All of them are captured in an identical scene with linear camera motion and collected for 10k samples. The only differences among them are shader setting: \textit{Raw} is rendered by no shader, \textit{middle-sildurs} uses sildurs with middle performance and \textit{high-sildurs} uses high-performance shader. We apply random initial weights encoder to monodepth2 and train it on above three datasets.
We use cross-evaluation on each trained model, i.e., evaluate every model on all datasets. The qualitative results are shown in \tab\ref{tab:shaders}.

\begin{table}[t!]
    \caption{The performance of the accuracy of the generated dataset under different rendering shaders. Here \textit{AbsRel} and $\delta<1.25^1$ are used as error and accuracy indicators. The best result in each row has been underlined and the optimal result has been bolded.}
    \label{tab:shaders}
    \centering
    \scalebox{0.9}{
    \begin{tabular}{l|ccc}
       \toprule[2pt] 
      & \multicolumn{3}{c}{\centering{Test Datasets}}\\
      \cline{3-3}
      \begin{tabular}[c]{@{}c@{}}Train Sets\\ (AbsRel $\backslash\delta^1$) \end{tabular}
      & \textit{raw} &\textit{middle-sildurs}&\textit{high-sildurs}  \\
      \hline
      \textit{raw}       &\underline{0.207$\backslash$0.689}   &0.326$\backslash$0.524         &0.311$\backslash$0.528\\
      \textit{middle-sildurs}  &0.436$\backslash$0.425         &\underline{0.148$\backslash$0.813}&0.158$\backslash$0.774\\
      \textit{high-sildurs}&0.439$\backslash$0.430         &0.156$\backslash$0.778        &\underline{\textbf{0.143$\backslash$0.816}}\\
      \toprule[2pt]
    \end{tabular}
    }
\end{table}

It shows that as the training scenes rendered gradually improve, the performance of the depth estimation model improves consequently.
Besides, compared with a model that is trained on less-texture data and tested on rendered data, the model that is trained on rendered data and tested on less-texture data brings a worse result.
It is consistent with the fact that the rendering performance will promote the robust of the model during the training.

\subsubsection{Lighting Conditions}
Previous study\cite{monodepth2,wxt2020} show that during the depth estimation model training, the low-texture areas caused by insufficient lighting or overexposure will produce problem pixels in depth estimation.
\begin{figure*}[t!]
    \begin{center}
        \includegraphics[width=1.0\linewidth]{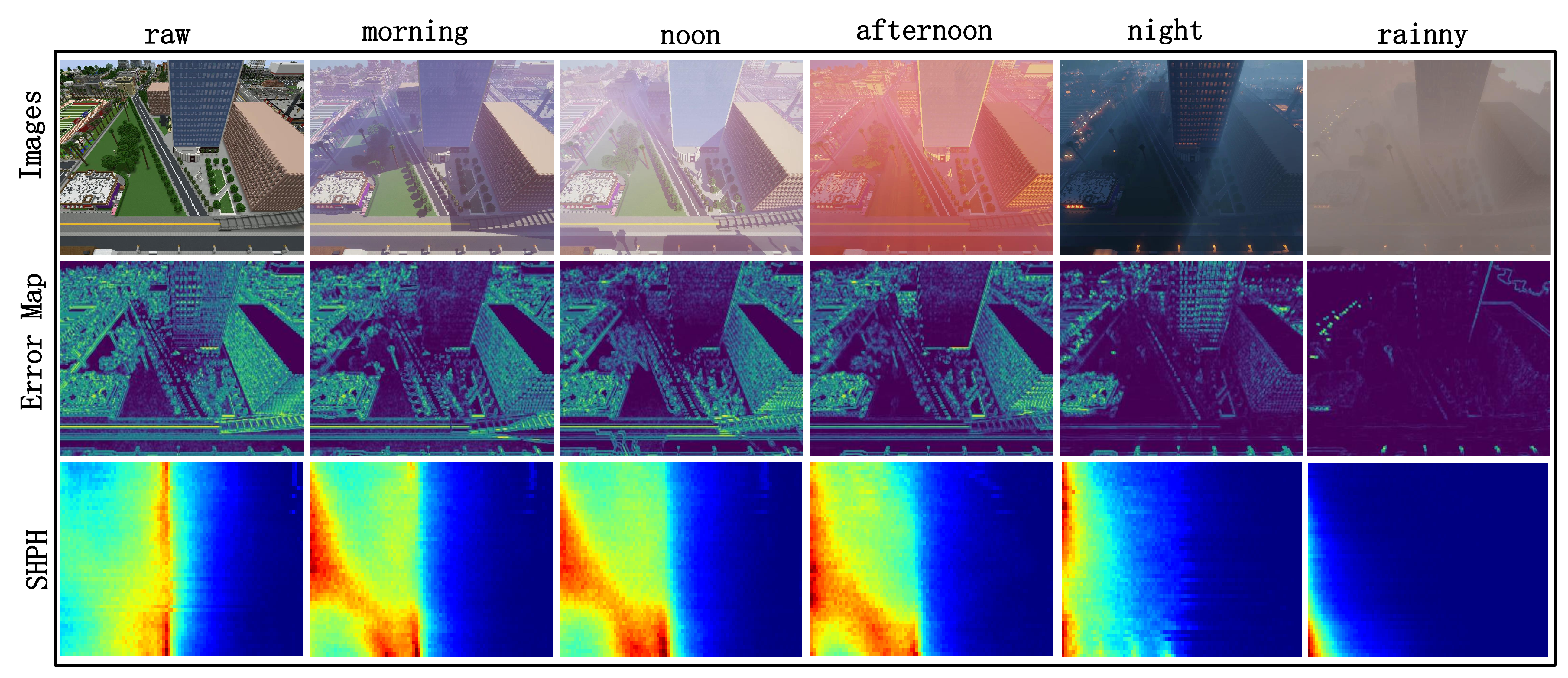}
    \end{center}
    \caption{Sequence under different lighting conditions. Photometric error map (row2) and SHEH map (row3).}  
        \label{fig:lights-show}
  \end{figure*}
\begin{figure}[t!]
    \begin{center}
        \includegraphics[width=0.65\linewidth]{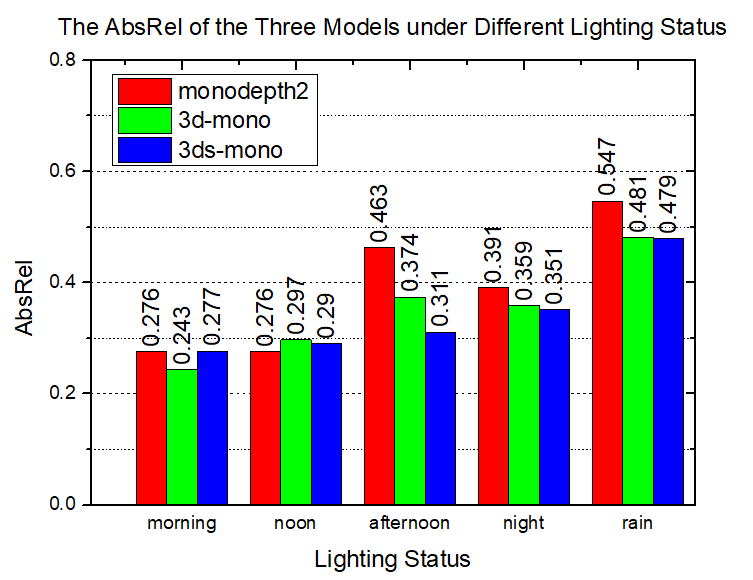}
    \end{center}
    \caption{Models trained with datasets that various in lighting conditions show different \textit{AbsRel}.}
        \label{fig:lights-colums}
\end{figure}

\begin{table*}[t!]
  \caption{Quantity Results in MineNavi with different lights} 
  \label{tab:lights}
  \centering
  \scalebox{1.0}{
  \begin{tabular}{l|l|cccc|ccc}
     \toprule[2pt] 
    && \multicolumn{4}{c|}{Error $\downarrow$} & \multicolumn{3}{c}{Accuracy $\uparrow$}  \\
    \cline{3-9}
    Models& dataset & AbsRel & SqRel & RMS & RMSlog & $<1.25$ & $<1.25^2$ & $<1.25^3$ \\\hline
    
    \multirow{4}{*}{monodepth2}
    & morning& 0.276 & 8.491 & 36.672 & 0.289 & 0.694 & 0.885 & 0.947 \\

    &noon & 0.276 & 11.284 & 49.412 & 0.321 & 0.665 & 0.858 & 0.931 \\
    
    &afternoon & 0.463 & 26.176 & 67.480 & 0.534 & 0.375 & 0.647 & 0.803 \\
    
    &night & 0.391 & 18.674 & 57.397 & 0.432 & 0.470 & 0.748 & 0.875 \\
    
    &rain  & 0.547 & 30.597 & 54.880 & 0.520 & 0.395 & 0.671 & 0.818 \\

    \toprule[0.5pt]
    \multirow{4}{*}{monodepth2-3D}
   & morning & 0.243 & 7.469 & 36.867 & 0.263 & 0.737 & 0.903 & 0.957 \\
&noon & 0.297 & 11.108 & 46.373 & 0.324 & 0.622 & 0.852 & 0.936 \\

&afternoon & 0.374 & 13.967 & 45.221 & 0.391 & 0.478 & 0.779 & 0.910 \\

&night & 0.359 & 15.065 & 49.514 & 0.392 & 0.490 & 0.780 & 0.907 \\

&rain  & 0.481 & 24.776 & 57.464 & 0.492 & 0.386 & 0.673 & 0.836 \\
\toprule[0.5pt]

\multirow{4}{*}{monodepth2-3Ds}
&morning & 0.277 & 10.086 & 41.467 & 0.290 & 0.697 & 0.883 & 0.950 \\

&noon& 0.290 & 11.167 & 46.460 & 0.315 & 0.656 & 0.865 & 0.936 \\

&afternoon & 0.311 & 11.901 & 43.584 & 0.322 & 0.618 & 0.859 & 0.942 \\

&night & 0.351 & 14.853 & 52.417 & 0.384 & 0.507 & 0.790 & 0.906 \\

&rain & 0.479 & 26.123 & 67.677 & 0.535 & 0.349 & 0.621 & 0.793 \\

    \toprule[2pt]
  \end{tabular}
  }
\end{table*}

To further explore the impact of lighting conditions in data to the depth estimation model, we apply the models  with random initial weights and train them on five datasets (See \fig\ref{fig:lights-show}) under different lighting conditions: morning, noon, afternoon, night and rainy day.
Quantified results on \textit{ArsRel} are shown in \fig\ref{fig:lights-colums}.
 We can observe that in the lighting conditions at morning and noon, three test models achieve similar results. However, as the lighting in training data is getting dim (afternoon, night, rainy), three models are deteriorated significantly. This can be attributed primarily to that the adequate lighting makes the color between pixels more diverse, and the error map is close to the uniform distribution.
Note that at the time of \textit{afternoon}, the models performance dropped dramatically, even worse than \textit{night} that with dimmer lighting condition, we suspect that the reason behind this is there are too much problematic pixels in captured images caused by lens flare, which strongest in \textit{afternoon} compared with the other lighting conditions.
SHPH results on the collected sequences under different lighting conditions and different camera moving paths are shown in the row3 of \fig\ref{fig:lights-show}. It can be clearly seen that the clear lighting conditions bring the even distribution of the SHPH.

\subsubsection{Impact of Motion Blur}

\begin{figure}[t!]
    \begin{center}
        \includegraphics[width=0.85\linewidth]{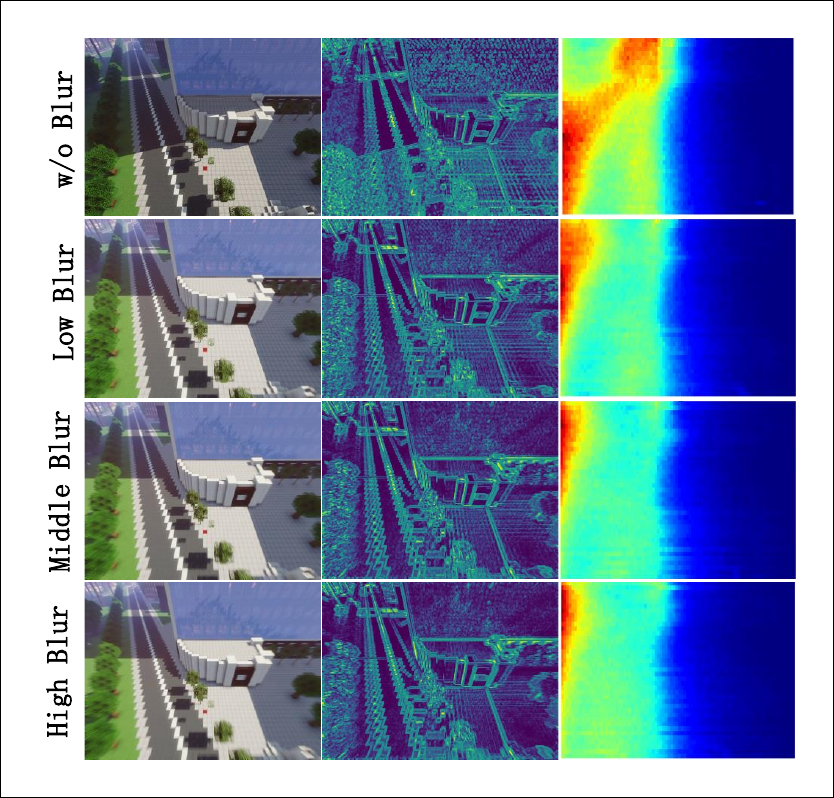}
    \end{center}
    \caption{At close to the ground ($ \leq 70m$) a histogram statistical result of the inter-frame error and its sequence photometric error heat map}
        \label{fig:mb-lines}
  \end{figure}

\begin{table}[t!]
    \caption{The model shows the different accuracy results under different motion blurs}
    \label{tab:motion-blur}
    \centering
    \scalebox{0.9}{
    \begin{tabular}{l|cccc}
       \toprule[2pt] 
      & \multicolumn{3}{c}{\centering{Test Datasets}}\\
      \cline{3-3}
      \begin{tabular}[c]{@{}c@{}}Train Sets\\ (\textit{AbsRel} $\backslash\delta^1$) \end{tabular}
      & \textit{None} &\textit{low blur}&\textit{middle blur}&\textit{high blur}  \\
      \hline
      \textit{None}       &\underline{0.221$\backslash$0.731} &0.232$\backslash$0.705 &0.235$\backslash$0.703 &0.237$\backslash$0.704\\
      \textit{low blur}   &0.237$\backslash$0.676 &0.203$\backslash$0.731 &\underline{0.199$\backslash$0.748} &0.201$\backslash$0.752\\
      \textit{middle blur}&0.203$\backslash$0.746 &0.177$\backslash$0.782 &\underline{\textbf{0.174$\backslash$0.811}} &0.179$\backslash$0.808 \\
      \textit{high blur}  &0.253$\backslash$0.646 &0.213$\backslash$0.692 &0.197$\backslash$0.729 &\underline{0.191$\backslash$0.754} \\

      \toprule[2pt]
    \end{tabular}
    }
\end{table}

The motions of cameras will also affect the stability of the SHPH. As shown in the \fig\ref{fig:mb-lines}, it can be seen that the distribution of the photometric error map gradually even  with the increase motion blur. In our experiment, four datasets with different motion blur are built.  The quantified results of monodepth2 are shown in the \tab\ref{tab:motion-blur}.
The motion blur has a great impact on the SHPH, we suspect that it is an effective way to overcome the noise and introduce the robustness by adding a certain motion blur in sequences. This is reflected in SHPH that appropriate motion blur can make the SHPH more stable, which leads the view synthesis of depth estimation model easer (see \fig\ref{fig:mb-lines}).

\begin{table}[t!]
  \caption{Motion blur test in monodepth2-3D (up) and monodepth2-3Ds (down). The best result in each row is underlined and the optimal result is bolded.}
  \label{tab:motion-blur-add}
  \centering

  \scalebox{0.8}{

    \begin{tabular}{l|cccc}
      \toprule[2pt] 
     & \multicolumn{3}{c}{\centering{Test Datasets}}\\
     \cline{3-3}
     \begin{tabular}[c]{@{}c@{}}Train Sets\\ (\textit{AbsRel} $\backslash\delta^1$) \end{tabular}
     & \textit{None} &\textit{low blur}&\textit{middle blur}&\textit{high blur}  \\
     \hline
     \textit{None}       &\underline{0.199$\backslash$0.768} &0.216$\backslash$0.760 &0.217$\backslash$0.758 &0.218$\backslash$0.752\\
     \textit{low blur}   &0.200$\backslash$0.730 &0.185$\backslash$0.764&\underline{0.184$\backslash$0.765} &0.186$\backslash$0.764\\
     \textit{middle blur}&0.194$\backslash$0.734 &0.177$\backslash$0.775 &\underline{\textbf{0.175$\backslash$0.781}}  &0.176$\backslash$0.785 \\
     \textit{high blur}  &0.200$\backslash$0.730 &0.186$\backslash$0.763 &\underline{0.180$\backslash$0.774} &0.181$\backslash$0.778 \\
  
     \toprule[2pt]
   \end{tabular}

    }

  \scalebox{0.8}{

  \begin{tabular}{l|cccc}
     \toprule[2pt] 
    & \multicolumn{3}{c}{\centering{Test Datasets}}\\
    \cline{3-3}
    \begin{tabular}[c]{@{}c@{}}Train Sets\\ (\textit{AbsRel} $\backslash\delta^1$) \end{tabular}
    & \textit{None} &\textit{low blur}&\textit{middle blur}&\textit{high blur}  \\
    \hline
    \textit{None}       &\underline{0.219$\backslash$0.739} &0.231$\backslash$0.738 &0.235$\backslash$0.737 &0.233$\backslash$0.735\\
    \textit{low blur}   &0.223$\backslash$0.704 &0.207$\backslash$0.734&0.206$\backslash$0.736 &\underline{\textbf{0.205$\backslash$0.737}}\\
    \textit{middle blur}&0.228$\backslash$0.691 &0.208$\backslash$0.726 &0.207$\backslash$0.731  &\underline{0.207$\backslash$0.734} \\
    \textit{high blur}  &0.246$\backslash$0.669 &0.226$\backslash$0.704 &0.224$\backslash$0.706 &\underline{0.223$\backslash$0.707} \\

    \toprule[2pt]
  \end{tabular}
  }

\end{table}

Table \ref{tab:motion-blur-add} shows the the performance of two variants of monodepth2 in MineNavi datasets with different motion blur.
It can be seen that the two models are trained on the motion-blurred dataset, and the performance is significantly better than the dataset without blurred.

Besides, we also introduce vary lighting conditions into experiments. As shown in the Table \ref{tab:lights}, it can be seen that in the variant models of monodepth2, the darker the lighting conditions, the worse the performance, which is consistent with the results of the experiments.
Note that at the time of \textit{afternoon}, the models performance dropped dramatically, even worse than \textit{night} that with dimmer lighting condition, We suspect that the reason behind this is there are too much problematic pixels in captured images caused by lens flare, which strongest in \textit{afternoon} compared with the other lighting conditions. It also suggest that the scene fidelity of MineNavi. 

\subsubsection{Impact of Ego-motion Variance}

\begin{table}[t!]
    \caption{Model performance in different ego-motion modes.
    } 
    \label{tab:motions}
    \centering
    \scalebox{0.9}{
    \begin{tabular}{l|ccc}
       \toprule[2pt] 
      & \multicolumn{3}{c}{\centering{velocities}}\\
      \cline{3-3}
      \begin{tabular}[c]{@{}c@{}}Train$\backslash$Test Sets\\ (AbsRel $\backslash\delta^1$) \end{tabular}
      & $v_1$ & $v_2$&$v_3$  \\
      \hline
     \quad $\lambda_1 = 1$ &\underline{0.143$\backslash$0.816}&\underline{\textbf{0.141$\backslash$0.818}}     & \underline{0.152$\backslash$0.806}\\
     \quad $\lambda_2 = \frac{\sqrt{2}}{2}$&0.240$\backslash$0.472   &0.240$\backslash$0.476         &0.252$\backslash$0.458  \\
     \quad $\lambda_3 = 0$&0.280$\backslash$0.414 &    0.473$\backslash$0.234   &0.481$\backslash$0.248\\
      \toprule[2pt]
    \end{tabular}
    }
\end{table}

The ego-motion of the camera in video will affect the depth estimation model training. Due to the continuous nature of the camera ego-motion, it is not easy to explore the impact of this factor. In this section, we build three datasets that various in motion mode which corresponds to linear motion $\lambda_1=1$ , overhead cruising motion$\lambda_2 = \frac{\sqrt{2}}{2}$, and circular motion  $\lambda_3 =0$ . 
Finally, the motion speed can be controlled by the number of interval frames of each train triplets in the datasets, and each of them are equipped with three velocity $v$, and $v_3 \textgreater v_2 \textgreater v_1$.
We test different motion modes through the models, and the quantitative results are shown in the \tab\ref{tab:motions}.
It can be seen that as the $\lambda$ decreases, the performance of the test model also decreases, and the velocity of training triplets also has significantly affect on the performance of test model. 
According to the previous analysis, the reason behind this probably is that the training triplet with larger $\lambda$ and appropriate velocity have a even distribution in SHPH, hence a better performance is achieved.

\subsubsection{Velocity of Training Sequence}

\begin{figure}[t!]
    \begin{center}
        \includegraphics[width=1.0\linewidth]{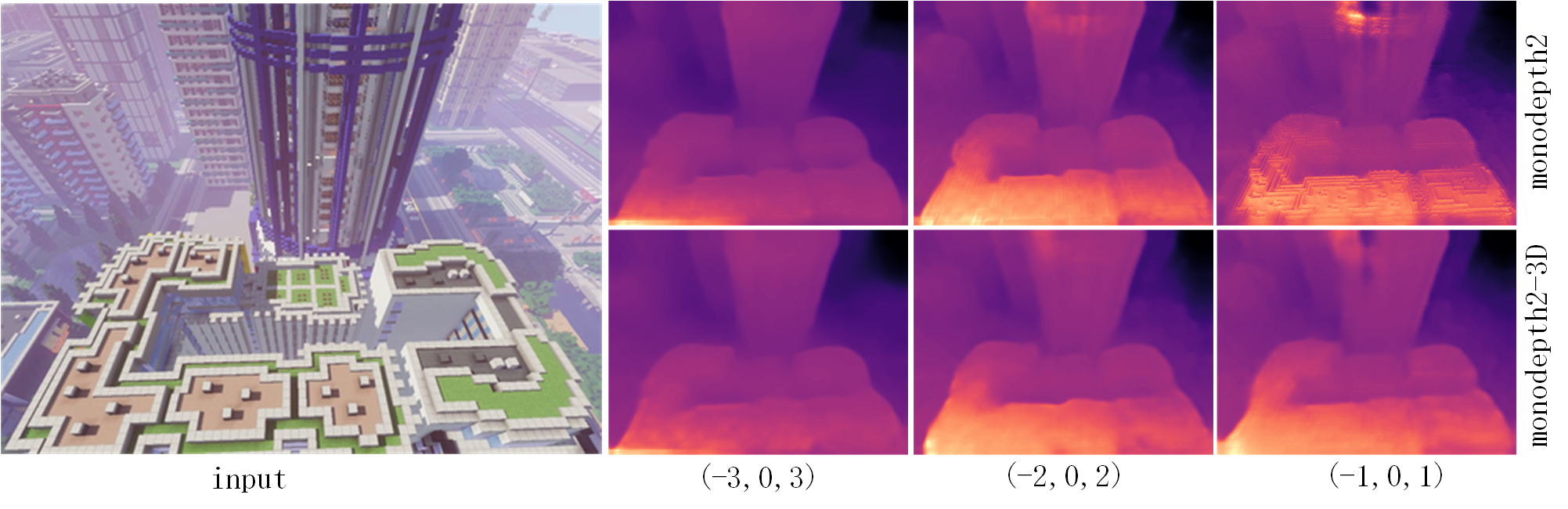}
    \end{center}
    \caption{Qualitative Model Results in MineNavi. First row: monodepth2. Second row: monodepth2-3D. From left to right, the velocity of training sequence is getting lower.}

        \label{fig:v-encoder-show}
  \end{figure}

We find that the training sequence vary in sample frequency can greatly affect the performance of the model. It is essential because if the velocity of sampling camera is faster, the photometric differences between two adjacent frames are bigger, making the model difficult to train.
\fig\ref{fig:v-encoder-show} shows the qualitative results of the models that vary in velocity of training sequence and encoder.

\section{Conclusion}
\label{sec:con}

This paper proposes a method to construct a synthetic dataset, which includes a large-scale scene with low cost but infinite volume, including surface normals, depth, and the 6DoF paths of the camera's ego-motion. 
This dataset generation method can provide a solution to overcome the difficulty of data collection in some dense estimation tasks. 
For depth estimation task in aircraft navigation, we construct several datasets. According to the experimental results, our proposed dataset generation method can perform as an intermediate domain for depth estimation. The data-to-model experiments reveal that future work should not only focus on the innovation of the models, but also pay more attention to the factors in the dataset that affect the models.




\ifCLASSOPTIONcaptionsoff
  \newpage
\fi

\bibliographystyle{IEEEtran}

\bibliography{ref}

\end{document}